# GR-Diffusion: 3D Gaussian Representation Meets Diffusion in Whole-Body PET Reconstruction

Mengxiao Geng, Zijie Chen, Ran Hong, Bingxuan Li, Qiegen Liu, *Senior Member, IEEE*

*Abstract*—Positron emission tomography (PET) reconstruction is a critical challenge in molecular imaging, often hampered by noise amplification, structural blurring, and detail loss due to sparse sampling and the ill-posed nature of inverse problems. The three-dimensional discrete Gaussian representation (GR), which efficiently encodes 3D scenes using parameterized discrete Gaussian distributions, has shown promise in computer vision. In this work, we propose a novel GR-Diffusion framework that synergistically integrates the geometric priors of GR with the generative power of diffusion models for 3D low-dose whole-body PET reconstruction. GR-Diffusion employs GR to generate a reference 3D PET image from projection data, establishing a physically grounded and structurally explicit benchmark that overcomes the low-pass limitations of conventional point-based or voxel-based methods. This reference image serves as a dual guide during the diffusion process, ensuring both global consistency and local accuracy. Specifically, we employ a hierarchical guidance mechanism based on the GR reference. Fine-grained guidance leverages differences to refine local details, while coarse-grained guidance uses multi-scale difference maps to correct deviations. This strategy allows the diffusion model to sequentially integrate the strong geometric prior from GR and recover sub-voxel information. Experimental results on the *UDPET* and *Clinical* datasets with varying dose levels show that GR-Diffusion outperforms state-of-the-art methods in enhancing 3D whole-body PET image quality and preserving physiological details.

## I. INTRODUCTION

Positron emission tomography (PET) is a cornerstone of molecular imaging, enabling non-invasive quantification of metabolic activity and physiological processes in oncology, neurology, and cardiology [1-3]. However, achieving high-quality PET reconstruction remains a significant challenge due to the inherently ill-posed nature of the inverse problem, which is exacerbated by factors such as low photon counts, limited scanning time, and the desire to minimize patient radiation exposure through low-dose protocols [4, 5]. These conditions lead to reconstructed images suffering from severe noise, streaking artifacts, and the loss of fine anatomical and pathological details, ultimately compromising diagnostic accuracy [6, 7].

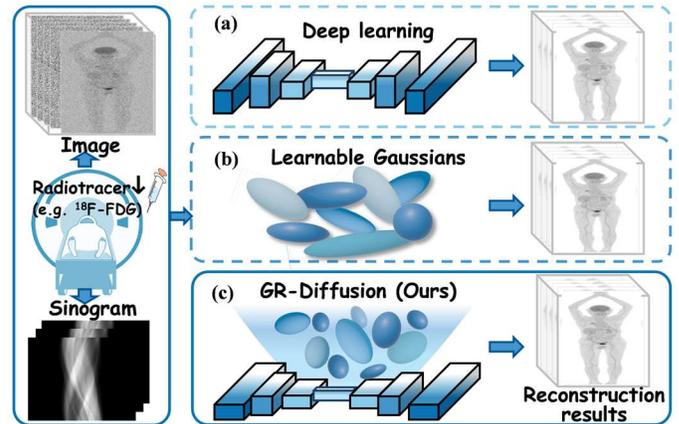

Fig. 1. Comparison among three kinds of PET semantic reconstruction methods: (a) deep learning, (b) learnable Gaussians, and (c) the proposed GR-Diffusion integrating both schemes.

Traditional methods like filtered back-projection (FBP) [8] are computationally efficient but prone to artifacts under low photon counts. Iterative methods such as ordered subset expectation maximization (OSEM) [9] further suppress noise and improve contrast by incorporating system models and regularization priors like total variation [10]. However, they still struggle to balance denoising with the preservation of structural details, often losing critical information due to over-smoothing [11]. The advent of deep learning has revolutionized the field of medical image reconstruction [12-15]. Deep learning methods represented by U-Net [16] have improved image quality, yet their results are prone to over-smoothing and suffer from limited generalization due to a lack of explicit physical modeling [17, 18]. While generative adversarial networks (GANs) [19, 20] can enhance visual realism, they face challenges such as training instability. More recently, diffusion models [21, 22] have emerged as a powerful solution for low-dose PET reconstruction. By learning to reverse a progressive noising process, these models can iteratively reconstruct high-fidelity images from noisy or low-quality inputs. In PET imaging, conditioning the reverse generation process on low-dose projection data enables diffusion models to demonstrate considerable potential in recovering fine structures and suppressing noise, thereby enhancing the visual quality and detail fidelity of the reconstructed images [23, 24]. Nevertheless, the generative process remains heavily reliant on prior distributions learned from large-scale datasets, making it difficult to strictly integrate the physical forward model of PET imaging as a hard constraint

This work was supported in part by the National Key Research and Development Program of China under Grants 2023YFF1204300 and 2023YFF1204302, the National Natural Science Foundation of China under Grant 62122033, and the Key Research and Development Program of Jiangxi Province under Grant 20212BBE53001. (M. Geng and Z. Chen are co-first authors.) (Corresponding authors: B. Li and Q. Liu.)

M. Geng, Z. Chen, R. Hong and Q. Liu are with School of Information Engineering, Nanchang University, Nanchang 330031, China. (mxiaogeng @163.com, zijiechen@email.ncu.edu.cn, ranhong@email.ncu.edu.cn, liuqiegen@ncu.edu.cn)

B. Li is with Anhui Province Key Laboratory of Biomedical Imaging and Intelligent Processing, Institute of Artificial Intelligence, Hefei Comprehensive National Science Center, Hefei 230088, China. (libingxuan@iai.ustc.edu.cn)

[25, 26], which may compromise the physical consistency and reliability of the reconstruction results.

In parallel, 3D Gaussian splatting (3DGS) [27] has emerged as a groundbreaking framework that represents a scene with a set of anisotropic 3D Gaussians, each parameterized by position, covariance, opacity, and view-dependent appearance [28]. This approach enables real-time, high-quality novel view synthesis through differentiable rasterization [29-31]. The success of 3DGS has inspired rapid adaptations to computer vision, where accurate 3D reconstruction is critical but challenged by sparse data, noise, and physical constraints. For instance, in X-ray [32-34] and computed tomography (CT) [35], methods like $R^2$-Gaussian [36] rectify integration biases in standard 3DGS to enable high-fidelity volumetric reconstruction from sparse projections, while GR-Gaussian [37] incorporates graph-based relationships to mitigate artifacts in sparse-view CT. Similarly, for dynamic CT, approaches such as deformation-informed 3D Gaussians [38] and $X^2$-Gaussian [39] model continuous 4D volumes by learning time-dependent Gaussian deformations, eliminating the need for phase-binning in respiratory motion analysis. Beyond CT, radiative Gaussian representations have been extended to magnetic resonance imaging (MRI) [40], where three-dimensional discretized Gaussian representation (GR) [41] uses learnable Gaussians for personalized reconstruction from under-sampled data, and to cryo-electron microscopy (Cryo-EM), with GEM [42] achieving efficient protein density mapping. In ultrasound imaging, UltraG-Ray [43] and UltraGS [44] leverage Gaussian splatting with physics-based acoustic modeling for novel view synthesis, while methods like PMF-STGR [45] integrate spatiotemporal Gaussians for dynamic CBCT reconstruction without prior models. Further innovations include wavelet-based primitives like WIPES [46], which enhance frequency flexibility for high-frequency details. These methods explicitly model the 3D volume, offering a compact, editable, and geometrically intuitive representation.

In this work, we propose GR-Diffusion, a novel framework that synergistically integrates a Gaussian Representation (GR) with a Diffusion model for 3D PET reconstruction, whose core fusion concept is illustrated in Fig. 1. Our work is driven by the key insight that an explicit 3D Gaussian representation, derived from the projection data, can provide a robust, physically grounded initial estimate of the 3D activity distribution. This Gaussian-based reconstruction serves not as a final product, but as a powerful geometric prior to guide a subsequent diffusion-based refinement process. The core of GR-Diffusion is a hierarchical guidance mechanism. First, it provides fine-grained guidance by leveraging feature differences to refine local details and recover sub-voxel information, allowing the diffusion model to sequentially integrate the strong geometric prior. Second, the Gaussian-reconstructed image generates a coarse-grained guidance signal, using multi-scale difference maps to correct deviations and ensure global consistency with the measured projections.

The main contributions are summarized as follows:

● **Unified GR-Diffusion Framework for PET Reconstruction.** We introduce GR-Diffusion, the first framework that seamlessly integrates an explicit 3D GR with a diffusion model for 3D PET reconstruction. This synergy establishes a new paradigm coupling explicit geometry with generative refinement, overcoming the limitations of standalone methods.

● **Theory-Grounded Gradient Guidance Mechanism.** We design a novel gradient guidance strategy rooted in mathematical analysis, which proves that the GR-derived reference image can surrogate ground-truth gradients under uncertainty. This enables anatomically consistent guidance during diffusion sampling, ensuring physical plausibility.

● **Multi-Scale Hierarchical Refinement**. We propose a dual-path guidance with fine-grained and coarse-grained components, allowing sequential integration of geometric priors for detail recovery and noise suppression.

The rest of this paper is organized as follows: Section II details the proposed GR-Diffusion method. Section III presents the experimental setup, results, and ablation studies. Finally, Section IV concludes the work.

## II. METHOD

Positron emission tomography is a cornerstone of molecular imaging, but the requirement for radioactive tracers to generate high-quality images imposes radiation exposure on patients. Long-term or repeated examinations may accumulate cancer risks and raise concerns about imaging safety. Thus, low-dose PET reconstruction must minimize the tracer dose while ensuring no loss of diagnostic information. In recent years, diffusion models have demonstrated strong potential in low-dose PET reconstruction by learning complex distribution priors from large-scale, high-quality image datasets. These methods can generate visually realistic and detail-rich images, but their performance is highly dependent on the size and quality of the training data. In real clinical scenarios, high-quality full-dose PET data are often scarce, and the noise distribution is complex with diverse degradation patterns. This leads to issues such as edge blurring, detail loss, or quantitative bias in purely data-driven diffusion models, limiting their clinical robustness. In contrast, 3D GR-based methods can learn distributions from raw sinogram data without depending on specific datasets, thereby avoiding reconstruction biases. By explicitly encoding 3D scenes using discretized Gaussian basis functions, GR methods stably extract subject-specific anatomical structure information, avoiding dependence on large datasets while maintaining strong geometric consistency under complex noise conditions. This dual advantage of dataset independence and anatomical fidelity positions GR methods as complementary alternatives to conventional data-driven approaches.

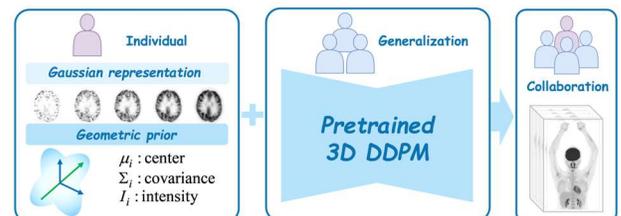

**Fig. 2.** Caption illustration of the proposed GR-Diffusion. It integrates individual Gaussian-based anatomical representations and generative refinement via a pretrained 3D diffusion, enabling collaborative subject-specific reconstruction.

The proposed GR-Diffusion framework is designed to reconstruct high-quality 3D whole-body PET images from low-dose projection data by synergistically integrating the subject-specific geometric priors with a generative refinement process. As illustrated in Fig. 2, the framework consists of two core components. On one hand, a 3D Gaussian representation module that constructs a physically grounded reference image from the raw sinogram data. On the other hand, a gradient-guided diffusion model that utilizes the GR output as a geometric prior to guide a reverse diffusion process for detail recovery and noise suppression. The following sections detail the formulation and implementation of each component.

### A. 3D Gaussian Representation for Guidance

The first stage of GR-Diffusion aims to transform the ill-posed, noisy projection data into an initial 3D reconstruction that possesses robust structural coherence. We achieve this by leveraging a 3D discrete Gaussian representation, a scene representation technique that models a 3D volume as a set of discrete, parameterized isotropic 3D Gaussian primitives. In Fig. 3, our methodology is organized into three key components: discretized Gaussian representation tailored for PET, fast volume reconstruction, and global optimization within the PET data acquisition model.

***Step 1 (Discretized Gaussian Representation for PET):*** In PET, the expected number of coincident events (sinogram data) $y$ is related to the underlying radiotracer activity image $\lambda$ through the following statistical model:

$$E[y] = P\lambda + r + s, \quad (1)$$

where $P$ is the system matrix, encoding the probability of an emission from a voxel being detected in a particular sinogram bin, $r$ is the expectation of random coincidences, and $s$ is the expectation of scattered events. The goal of low-dose PET reconstruction is to accurately estimate $\lambda$ from a low-dose sinogram $y$, which is an ill-posed inverse problem exacerbated by high Poisson noise.

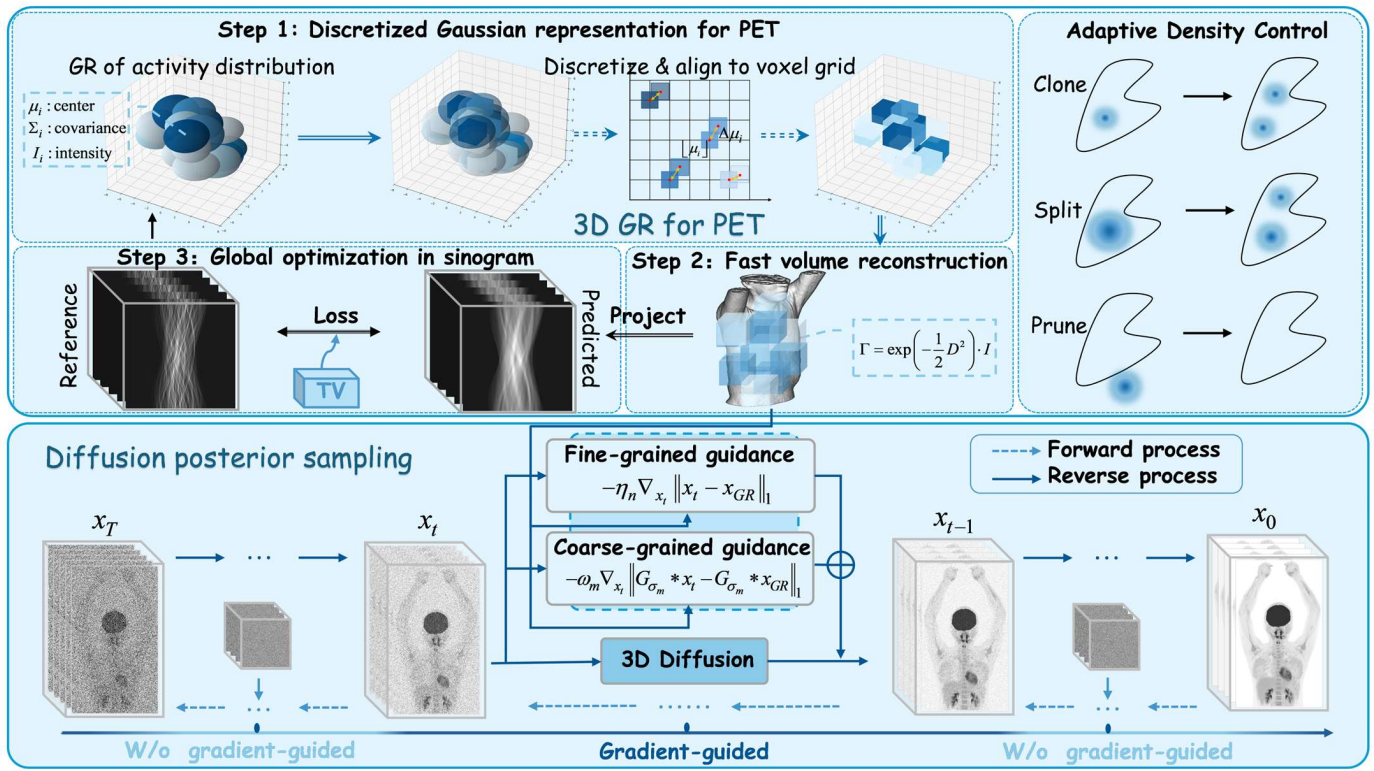

Fig. 3. Overview of the GR-Diffusion framework. This framework contains two key components: (1) a 3D Gaussian representation module that constructs a physically grounded reference image from raw sinogram data; (2) a gradient-guided diffusion model that takes the GR output as geometric prior to facilitate detail recovery and noise suppression in reverse diffusion.

***(1) GR of Activity Distribution:*** Inspired by the fact that radiotracer uptake in tissues often presents localized, blob-like distributions, we model the 3D activity image $\lambda$ as a linear combination of discrete, isotropic 3D Gaussian functions. Each Gaussian $G_i$ is defined by its center $\mu_i$, covariance matrix $\Sigma_i$ (controlled by a standard deviation $\sigma_i$), and a positive intensity $I_i$, which corresponds to the local activity concentration. The intensity of a voxel at position $p$ is given by:

$$\lambda(p) = \sum_{i=1}^{n} G(p, \mu_i, \Sigma_i) \cdot I_i$$
$$= \sum_{i=1}^{n} \exp(-\frac{1}{2}(p-\mu_i)^T \Sigma_i^{-1}(p-\mu_i)) \cdot I_i. \quad (2)$$

We employ isotropic Gaussians primarily for computational efficiency and to avoid over-parameterization. Given the ill-posed nature of low-dose PET reconstruction and the inherently noisy data, an isotropic representation provides a sufficient yet compact basis for modeling the activity distribution, leading to more stable optimization compared to anisotropic alternatives.

***(2) Discretize & Align to Voxel Grid:*** To ensure differentiability and precise alignment with the discrete voxel grid, we employ the "discretize and align" procedure [37]. The continuous Gaussian center $\mu_i$ is decomposed into its floored integer coordinates $\lfloor \mu_i \rfloor$ and a fractional offset $\Delta \mu_i = \mu_i - \lfloor \mu_i \rfloor$. The Gaussian's contribution is then computed within a local cuboid region $\lfloor B \rfloor$ centered at $\lfloor \mu_i \rfloor$, with the coordinates adjusted by $\Delta \mu_i$ to maintain accurate spatial alignment. This allows gradients to flow back to the continuous parameters $\mu_i$ and $\sigma_i$ during optimization.

*Step 2 (Fast Volume Reconstruction):* Directly computing the contribution of every Gaussian to every voxel in the volume is computationally prohibitive. We therefore confine the effect of each Gaussian to a local region around its center (e.g., a 11×11×11 voxel box). Our highly parallelized fast volume reconstruction technique efficiently aggregates these local contributions. The core idea lies in the decomposition of the squared Mahalanobis distance $D^2$ ($D^2 = \lfloor B \rfloor^T C^{-1} \lfloor B \rfloor$) calculation. Instead of a monolithic computation, we break it down into pre-computable components involving the fixed local coordinate box $B$ and the per-Gaussian offset $\Delta \mu$:

$$S_{BB} = B^T C^{-1} B, \quad S_{B\Delta\mu} = B^T C^{-1} \Delta\mu, \\ S_{\Delta\mu B} = \Delta\mu^T C^{-1} B, \quad S_{\Delta\mu\Delta\mu} = \Delta\mu^T C^{-1} \Delta\mu. \quad (3)$$

The squared Mahalanobis distance $D^2$ for the aligned discretized coordinates $\lfloor B \rfloor$ is then efficiently computed as a linear combination:

$$D^2 = S_{BB} - S_{B\Delta\mu} - S_{\Delta\mu B} + S_{\Delta\mu\Delta\mu}. \quad (4)$$

This decomposition drastically reduces the computational overhead. The final Gaussian contribution $\Gamma$ to each voxel in the local region is calculated as $\Gamma = \exp(-\frac{1}{2}D^2) \cdot I$. These contributions are then summed into the corresponding voxels of the 3D activity image $\lambda$ in a parallelized scatter-add operation.

*Step 3 (Global Optimization in Sinogram):* After reconstructing the activity image $\lambda$ from the Gaussians, we optimize the Gaussian parameters $\{\mu_i, \sigma_i, I_i\}$ by minimizing the discrepancy between the forward-projection of $\lambda$ and the measured low-dose sinogram $y$.

*(1) Forward Projection and Loss Function:* The reconstructed activity image $\lambda$ is forward-projected using the system matrix $P$ to generate an estimated sinogram $\hat{y}$:

$$\hat{y} = P\lambda. \quad (5)$$

Given the Poisson nature of PET data, the optimization is guided by a loss function that combines a Poisson likelihood-based term (e.g., a weighted least squares or Kullback-Leibler divergence approximation) with regularization terms suitable for PET:

$$\mathcal{L}_{total} = \lambda_1 \mathcal{L}_{Data}(\hat{y}, y) + \lambda_2 \mathcal{L}_{TV}(\lambda), \quad (6)$$

where $\mathcal{L}_{Data}$ is the data fidelity term, such as the negative log-likelihood of the Poisson distribution. $\mathcal{L}_{TV}$ (total variation) encourages piecewise smoothness, a valid prior for PET images, helping to suppress noise while preserving edges.

*(2) Adaptive Density Control:* We incorporate an adaptive density control strategy [27] for the 3D Gaussians, which operates periodically based on analytic gradients. This process involves three core operations:

(i) Pruning: Gaussians with persistently low intensity gradients ($|\nabla_{I_i} \mathcal{L}_{total}| < \tau_{prune}$), indicating negligible contribution to the PET image, are removed.

(ii) Cloning: Areas with high reconstruction error ($|\nabla_{\mu_i} \mathcal{L}_{total}| < \tau_{clone}$), signaling under-representation, trigger the cloning of Gaussians to enhance local detail.

(iii) Splitting: Over-sized Gaussians ($\sigma_i < \tau_{split}$) are split into smaller units to improve spatial resolution.

### B. Gradient-Guided Diffusion with Geometric Prior

Although the first stage already yields satisfactory reconstruction performance, it fundamentally provides an approximate reconstruction of physical morphology and thus fails to meet the stringent requirements of PET imaging for metabolic quantification accuracy and fine textural detail. To address this limitation, we design a diffusion-based reconstruction framework guided by geometric GR priors, which performs perceptual refinement and detail enhancement on the physically grounded morphology produced by 3D GR. Specifically, we pre-train a 3D denoising diffusion probabilistic model (DDPM) on a large-scale dataset of high-quality whole-body PET images to serve as the base generative model. The reconstructed images by GR are then decomposed into fine-grained and coarse-grained components, which jointly guide the sampling process of the diffusion model. This approach enables the reconstruction of high-fidelity PET images that preserve both physical plausibility and quantitative consistency.

*Pretrained 3D DDPM:* We train a 3D DDPM using a large dataset of high quality full-dose PET images to capture the distribution of complex textures and precise quantitative values in high quality PET images. DDPM operates by progressively adding Gaussian noise to the clean volume $x_0$ through a Markov chain, governed by a predefined scale scheduler $\{\beta_t \in (0,1)\}_{t=1}^T$, the forward process at step $t$ is defined as:

$$q(x_t | x_{t-1}) = \mathcal{N}(x_t; \sqrt{1-\beta_t} x_{t-1}, \beta_t \mathcal{I}). \quad (7)$$

Let $\alpha_t = 1 - \beta_t$ and $\bar{\alpha}_t = \prod_{s=1}^t \alpha_s$, then the forward process can be expressed in closed form as a transition from the initial state $x_0$ to any intermediate state $x_t$:

$$q(x_t | x_0) = \mathcal{N}(x_t; \sqrt{\bar{\alpha}_t} x_0, (1-\bar{\alpha}_t)\mathcal{I}). \quad (8)$$

During the reverse process, the model estimates the clean volume conditioned on the noisy observation and the current time step $t$. The reverse transition distribution, derived from the forward process is formulated as:

$$p_\theta(x_{t-1} | x_t) = \mathcal{N}(x_{t-1}; \mu_\theta(x_t, t), \Sigma_\theta(x_t, t)), \quad (9)$$

where $\mu_\theta$ and $\Sigma_\theta$ denote the parameterized mean and variance of Gaussian distribution parameterized by the neural network, and $\theta$ represents the learnable parameters. The reverse process uses reparameterization to convert predicting the mean and variance into the predicting the noise $\varepsilon_\theta(x_t, t)$. Consequently, the reverse update can be expressed as:

$$x_{t-1} = \frac{1}{\sqrt{\alpha_t}} \left( x_t - \frac{\beta_t}{\sqrt{1-\bar{\alpha}_t}} \varepsilon_\theta(x_t, t) \right) + \sigma_t z, \quad (10)$$

where $z \sim \mathcal{N}(0,1)$.

We adopt a 3D-UNet architecture as the backbone network for noise prediction, comprising five down-sampling and five up-sampling stages, each equipped with two 3D convolutional layers without any attention blocks. For computational efficiency and memory consumption, during training, we randomly sample contiguous axial volumes of 96 slices from each 3D PET image as input $x_0$. During the iterative reconstruction process, we extract overlapping volumes of 96 slices along the axial direction, with an overlay of 16 slices between consecutive volumes to ensure spatial consistency. This strategy enables the model to capture accurate 3D structural features under constrained computational resources and supports high-fidelity image generation in subsequent refinement stages.

*Gradient-Guided Diffusion Process:* After pre-training the 3D DDPM, we leverage the images $x_{GR}$ reconstructed during the 3D-GR stage as geometric priors. We compute the distance between this prior and the diffusion reverse trajectory, then use the gradient of this distance as a guidance term to steer the reverse generation process of the diffusion model.

**Theorem 1.** Let $D = x_{gt} - x_{GR}$, $D_t = x_{gt} - x_t$, and assume $\|D\|_\infty < \xi$. Then, for any $t$ such that $\|D_t\| > \xi$, the following holds:

$$\nabla_{x_t} \|x_t - x_{GR}\|_1 \simeq \nabla_{x_t} \|x_t - x_{gt}\|_1. \quad (11)$$

*Proof.* We begin by noting the identity

$$\nabla_{x_t} \|x_t - x_{GR}\|_1 = \nabla_{x_t} \|D - D_t\|_1. \quad (12)$$

Since the $\ell_1$-norm is non-differentiable at zero, its sub-gradient is given by the sign function:

$$\nabla_{x_t} \|D - D_t\|_1 = \text{sign}(D - D_t), \quad (13)$$

where $\text{sign}(x) = 1$ if $x > 0$, and $\text{sign}(x) = -1$ otherwise.

Under the assumption that $\|D\|_\infty < \xi$ and $\|D_t\| > \xi$, we have $\|D_t\| > \|D\|$. This implies that the magnitude of $D_t$ dominates that of $D$, so

$$\text{sign}(D - D_t) = \text{sign}(-D_t). \quad (14)$$

On the other hand, $\nabla_{x_t} \|x_t - x_{gt}\|_1 = \text{sign}(x_t - x_{gt}) = \text{sign}(-D_t)$. Combining these results yields

$$\nabla_{x_t} \|x_t - x_{GR}\|_1 = \nabla_{x_t} \|x_t - x_{gt}\|_1, \quad (15)$$

which completes the proof. □

**Remark 1.** In the early stages of the reverse diffusion process, the discrepancy between $x_t$ and $x_{gt}$ is typically large. Under the condition $\|D_t\| > \xi$, the gradient $\nabla_{x_t} \|x_t - x_{GR}\|_1$ provides a valid surrogate for $\nabla_{x_t} \|x_t - x_{gt}\|_1$. Thus, this gradient can be employed to guide the pre-trained 3D DDPM during sampling, effectively substituting high quality supervision when direction access to $x_{gt}$ is unavailable.

Building upon Theorem 1 and as illustrated in Fig. 3, we propose a dual-path gradient guidance strategy that leverages the pre-trained 3D DDPM to refine both the structural detail and quantitative fidelity of GR-reconstructed images. Specifically, we incorporate the gradient of the $\ell_1$-distance between $x_{GR}$ and $x_t$ into the reverse diffusion process, thereby steering the generative trajectory toward solutions that preserve patho-logical consistency.

In the early phase of the reverse process, when image morphology and anatomical contours remain poorly defined and noise dominates the signal, the $\ell_1$-distance between $x_{gt}$ and $x_t$ may be effectively approximated by that between $x_{GR}$ and $x_t$. We refer to this strategy as fine-grained guidance. However, as the reverse process advances and $x_t$ converges toward $x_{gt}$, the discrepancy between $x_{GR}$ and $x_t$ becomes increasingly misleading, potentially diverting the denoising trajectory. To mitigate this, we introduce multi-scale Gaussian kernels to smooth the discrepancy between $D = x_{gt} - x_{GR}$ and $D_t = x_{gt} - x_t$, thereby suppressing spurious fine-scale gradients arising from $x_{GR}$. We term this regularized gradient signal coarse-grained guidance.

*(1) Fine-Grained Guidance:* In the case of fine-grained guidance, where $\|D_t\| > \xi$, as shown in Fig. 4, we use the $\ell_1$-norm to represent the distance between the two images. Based on Theorem 1, $\nabla_{x_t} \|x_t - x_{GR}\|_1 = \nabla_{x_t} \|x_t - x_{gt}\|_1$. Therefore, the conditional guidance term in the reverse process can be written as:

$$\nabla_{x_t} \log p(y | x_t) = -\eta_n \nabla_{x_t} \|x_t - x_{GR}\|_1 \quad (16)$$

where $x_t = \frac{1}{\sqrt{\alpha_t}} \left( x_t - \frac{\beta_t}{\sqrt{1-\bar{\alpha}_t}} \varepsilon_\theta(x_t, t) \right) + \sigma_t z$ and $\eta_n$ is used to adjust the strength of the gradient guidance, $y$ represents the reference condition, which in the case of fine-grained guidance is $x_{GR}$.

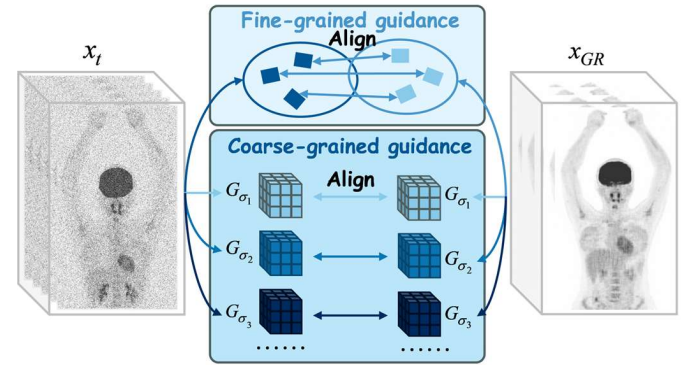

**Fig. 4.** Overview of the fine-grained and coarse-grained guidance processes.

*(2) Coarse-Grained Guidance:* As illustrated in Fig. 4, our coarse-grained guidance employs multi-scale Gaussian kernels to blur the predicted and reference images, thereby alleviating the detrimental influence of spurious signals from $x_{GR}$. The corresponding conditional guidance term in the reverse process is given by:

$$\nabla_{x_t} \log p(y | x_t) = -\omega_m \sum_{s=1}^{N} \delta_s \nabla_{x_t} \|G_{\sigma_s} * \hat{x}_t - G_{\sigma_s} * x_{GR}\|_1 \quad (17)$$

where $G_{\sigma_s}$ denotes a zero-mean Gaussian kernel with variance $\sigma_s$. $\delta_s$ weights the gradient contribution of each Gaussian kernel, while $\omega_m$ scales the overall strength of the coarse-grained guidance term. For coarse-grained guidance, the reference condition $y$ is $G_{\sigma_s} * x_{GR}$. In our implementation, we employ three Gaussian kernels, each with a kernel size of $3\times3\times3$ and variances of 1.0, 2.0, and 4.0, respectively.

*Guidance Schedule:* We apply dual-path gradient guidance for the partial reverse diffusion process. First, computing gradient guidance at every time step incurs substantial computational overhead. Second, we observe that gradient guidance has limited efficacy in the very early phase of the reverse process, where the image is dominated by high-amplitude noise; conversely, excessive gradient guidance in the final phase can disrupt the formation of fine image details. Thus, we activate both fine-grained and coarse-grained guidance only during the middle phase of the reverse process. Specifically, the reverse process runs from $T$ to 0, and we restrict guidance to the interval $[0.4T, 0.6T]$. This strategy avoids excessive gradient perturbation, provides adequate guidance to the DDPM, and introduces negligible computational overhead.

In summary, the pipelines that perform iterative reconstruction by combining 3D-GR with a pre-trained 3D DDPM are fully specified in **Algorithm 1.**

---

**Algorithm 1 GR-Diffusion**

**Iterative reconstruction**

1: **Input:** 3D DDPM $\varepsilon_\theta$, low-dose measurement $y$
2: **Rendering:** $x_{GR} = \sum_i^n G(p_i, \mu_i, \Sigma_i) \cdot I_i$
3:      $x_T \sim \mathcal{N}(0,1)$
4: **for** $t$ **in** $[T, T-1, \cdots, 1]$ **do**
5:      $x_{t-1} = \frac{1}{\sqrt{\alpha_t}}\left(x_t - \frac{\beta_t}{\sqrt{1-\bar{\alpha}_t}}\varepsilon_\theta(x_t, t)\right) + \sigma_t z$
6:      **if** $t \in [\text{st}, \text{ed}]$ **then**
7:          $g_f \leftarrow \nabla_{x_t} \log p(x_{GR}|x_t)$ **by Eq. (16)**
8:          $g_c \leftarrow \nabla_{x_t} \log p(G_{\sigma_s} * x_{GR}|x_t)$ **by Eq. (17)**
9:          $\hat{x}_{t-1} \leftarrow x_{t-1} + g_f + g_c$
10:     **else**
11:         $\hat{x}_{t-1} \leftarrow x_{t-1}$
12:     **end if**
13: **end for**
14: **Output:** Final reconstructed PET image $x_0$

---

## III. EXPERIMENTS

### A. Experimental Setup

In this section, the reconstruction performance of GR-Diffusion is compared with state-of-the-art methods, including UNet-2D [16], UNet-3D [17], DDPM-2D [21], DDPM-3D [47], and emerging 3D Gaussian-based methods: R²-Gaussian [36], X²-Gaussian [39], and DGR [41]. Notably, R²-Gaussian, X²-Gaussian and DGR represent the first implementation of 3D Gaussian representation in PET imaging reconstruction, which have previously demonstrated exceptional performance in computer vision and CT reconstruction. To ensure comparability and fairness of the experiments, all methods are conducted with the same computing resources and datasets. Open-source code is available at: https://github.com/yqx7150/GR-Diffusion.

*Datasets:* The *UDPET* dataset, used to train the 3D DDPM and evaluated the proposed methods, was obtained from the MICCAI 2024 Ultra-low Dose PET Imaging Challenge. It includes 377 groups of whole-body $^{18}$F-FDG PET scans, consisting of full dose and low dose scans with dose reduction factors (DRFs) of 4, 10, 20, 50 and 100. A total of 347 pairs of full-dose and low-dose PET images were used to train the 3D DDPM. Each patient's scan includes 673 axial 2D slices, with the resolution of 360×360. During the training phase, to reduce the consumption of computing resources, for each patient's scan, we randomly select 96 axial slices to form a 3D volume for the 3D DDPM to learn data distribution. Additionally, each axial slice is centrally cropped to a size of 192×288, aiming to retain only the scanned anatomy while eliminating the blank regions. The remaining 30 data samples were used for evaluation. During the evaluation phase of the GR-Diffusion, the volume, which fed to 3D DDPM, is generated in batches of 96 axial slices. To mitigate discrepancies between adjacent volumes, an overlap of 16 slices was introduced between two consecutive volumes. The *Clinical* dataset consists of whole-body PET scans acquired from 19 patients using the DigitMI 930 PET/CT scanner (developed by RAYSOLUTION Healthcare Co., Ltd.), which is equipped with all-digital PET detectors and features an axial field-of-view (AFOV) of 30.6cm within an 81cm ring diameter. Each patient underwent a scan covering 4 to 8 bed positions, with a complete sampling scan time ranging from 45 s to 3 min per bed. Low-dose PET data were generated through resampling at regular time intervals. Specifically, the list-mode data were segmented according to the chronological sequence, with each cycle corresponding to a 2 millisecond (ms) interval. Within each cycle, data acquired during a 1 ms interval were retained, while the remaining data were discarded. The resulting low-count data were then rearranged into the im-age domain. This study was approved by the institutional review board of Beijing Friendship Hospital, Capital Medical University, Beijing, China (Approval No. 2022-P2-314-01).

*Parameter Configuration:* For the 3D GR geometric prior, 200,000 Gaussian kernels with a spatial support size of 11 × 11 × 11 were employed to initialize the low-dose PET reconstruction, with the maximum number of Gaussian kernels capped at 250,000. During GR optimization, the reconstruction was performed for 3,000 iterations to obtain the final GR-reconstructed image. An adaptive density control strategy was applied during the GR reconstruction stage, where the pruning threshold was set to $\tau_{\text{prune}} = 1e-7$, and the cloning and splitting thresholds were set to $\tau_{\text{clone}} = 2e-4$ and $\tau_{\text{split}} = 0.01$, respectively, to dynamically adjust the distribution of Gaussian primitives. For the gradient-guided diffusion process, to avoid excessive computational overhead, only 3D patches with a size of 192×288×96 is generated per batch, and a half-precision inference strategy is adopted. To ensure reconstruction accuracy, during the reverse denoising process, the standard Euler method is employed without any accelerated sampling strategies. Additionally, we set the number of reverse denoising steps to 1000 and integrate both coarse-grained and fine-grained gradient guidance during the denoising step interval of 400 to

600. All training and evaluation experiments were conducted using two NVIDIA GeForce RTX 3090 GPUs, each with 24 GB of memory.

*Performance Evaluation:* To quantitatively assess the reconstruction quality of GR-Diffusion, the peak signal-to-noise ratio (PSNR), structural similarity (SSIM), mean squared error (MSE), and Fréchet inception distance (FID) are employed.

### B. Reconstruction Experiments

*Comparison on UDPET Public Dataset:* To assess the efficacy of GR-Diffusion using the *UDPET* dataset, Table I displays a comprehensive quantitative analysis across different DRFs. The results demonstrate that GR-Diffusion consistently achieves higher PSNR and SSIM values, while exhibiting lower MSE and FID values compared to other reconstruction methods. At DRF = 20, GR-Diffusion achieves the highest PSNR of 44.37 dB, outperforming the second-best DDPM-3D method by 0.48 dB, while simultaneously yielding the lowest reconstruction error with an MSE of 0.0094. When the dose reduction factor is further increased to DRF = 50, the advantage of GR-Diffusion becomes more pronounced. GR-Diffusion attains a PSNR of 43.53 dB, surpassing the 3D diffusion baseline DDPM-3D by 1.05 dB in PSNR. In addition, GR-Diffusion maintains the lowest reconstruction error with an MSE of 0.0113, indicating superior robustness under extremely low-dose conditions. These results demonstrate that GR-Diffusion consistently preserves both structural fidelity and perceptual image quality across varying dose reduction factors, highlighting its strong generalization capability and effectiveness for low-dose PET reconstruction.

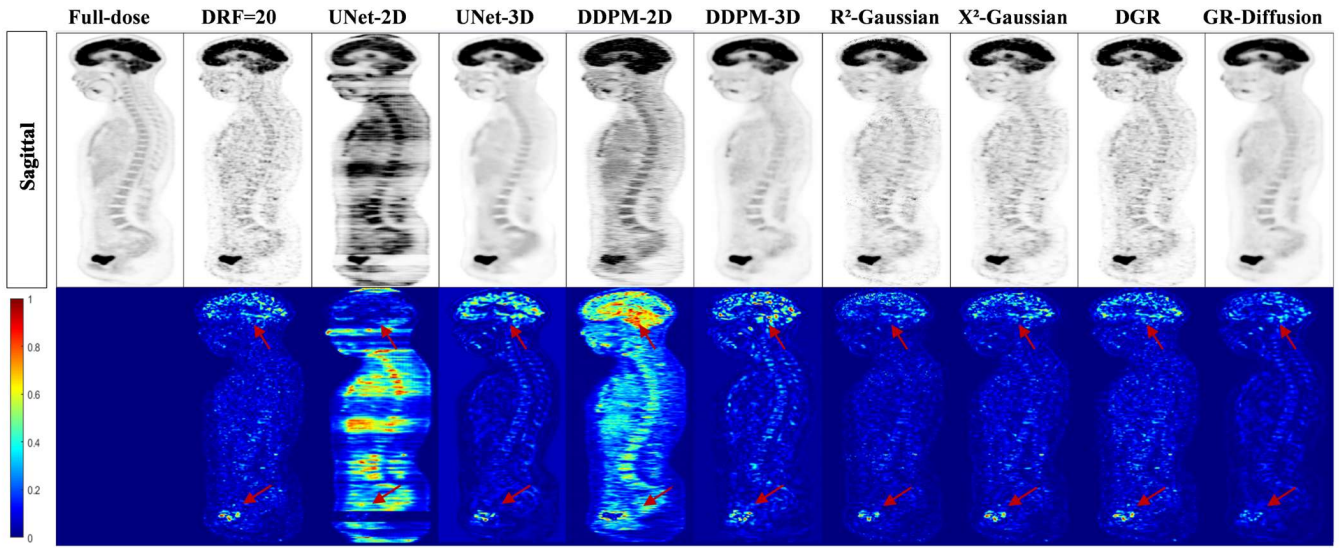

**Fig. 5.** Qualitative comparison of reconstruction results on the *UDPET* public dataset at a DRF value of 20. From left to right: Full-dose, low-dose, reconstruction by UNet-2D, UNet-3D, DDPM-2D, DDPM-3D, R²-Gaussian, X²-Gaussian, DGR, and GR-Diffusion. The corresponding error maps with respect to the full-dose reference are presented below each reconstructed image, where representative differences are highlighted by red arrows.

Fig. 5 presents a qualitative comparison of whole-body PET reconstruction results at DRF = 20 on the *UDPET* public dataset, including sagittal views with their corresponding error maps. Fig. 6 shows representative axial PET reconstruction results at two different anatomical levels on the UDPET public dataset under the DRF = 20 condition, together with their corresponding error maps.

From the sagittal reconstructions, it can be observed that UNet-2D and DDPM-2D suffer from evident inter-slice discontinuities, manifested as banding artifacts and structural breaks along the axial direction. These artifacts are particularly noticeable in the torso and lower limb regions, indicating limited capability of 2D-based models to enforce cross-slice consistency in volumetric reconstruction. However, in the axial views shown in Fig. 6, these two methods yield relatively coherent local structures, suggesting that their limitations mainly arise from the lack of explicit 3D contextual modeling rather than in-plane reconstruction accuracy.

In contrast, R²-Gaussian, X²-Gaussian, and DGR exhibit improved volumetric consistency in the sagittal views, producing smoother global anatomical layouts without obvious slice-wise discontinuities. Nevertheless, their reconstructions tend to retain noticeable noise and lose fine structural details, especially under low-dose conditions and in axial slices with complex uptake patterns, as reflected by fragmented intensity distributions and elevated local errors in the corresponding error maps. Among the remaining methods, UNet-3D and DDPM-3D alleviate inter-slice artifacts, yielding more continuous structures across views. UNet-3D effectively suppresses noise but introduces over-smoothing that blurs tissue boundaries, while DDPM-3D better preserves structural continuity yet still exhibits residual noise and localized distortions in regions with sharp intensity transitions.

GR-Diffusion consistently demonstrates superior reconstruction quality across all views. In both whole-body projections and axial slices, GR-Diffusion produces intensity distributions and anatomical structures that closely resemble the full-dose reference, while maintaining sharp boundaries and coherent tracer uptake patterns. The corresponding error maps show lower error magnitudes and a more homogeneous spatial distribution, indicating stable reconstruction behavior under severe dose reduction. Notably, GR-Diffusion effectively suppresses low-dose noise without introducing slice-wise artifacts or excessive smoothing, preserving both global anatomical consistency and local structural fidelity across different anatomical levels.

TABLE I
COMPARISON OF STATE-OF-THE-ART METHODS IN TERMS OF AVERAGE PSNR ↑, SSIM ↑, MSE (*E-3) ↓, AND FID ↓ UNDER VARIOUS DRFS ON THE *UDPET* AND *CLINICAL* DATASETS. THE **BOLD** AND *ITALIC* FONTS INDICATE THE OPTIMAL AND SUB-OPTIMAL VALUES, RESPECTIVELY.

| Dataset | DRF | Metric | UNet-2D | UNet-3D | DDPM-2D | DDPM-3D | $R^2$-Gaussian | $X^2$-Gaussian | DGR | GR-Diffusion |
|---|---|---|---|---|---|---|---|---|---|---|
| *UDPET* | 20 | PSNR↑ | 17.64 | 39.64 | 23.87 | *43.89* | 41.72 | 42.95 | 42.97 | **44.37** |
| | | SSIM↑ | 0.8278 | *0.9833* | 0.8558 | **0.9834** | 0.9689 | 0.9760 | 0.9763 | 0.9705 |
| | | MSE↓ | 4.4069 | 0.0278 | 1.0502 | *0.0105* | 0.0172 | 0.0130 | 0.0129 | **0.0094** |
| | | FID↓ | 162.9043 | *79.3234* | 88.4102 | 105.4330 | 144.1479 | 125.8885 | 119.4779 | **73.5345** |
| | 50 | PSNR↑ | 13.13 | 31.79 | 27.49 | *42.47* | 36.74 | 36.74 | 42.42 | **43.53** |
| | | SSIM↑ | 0.8732 | 0.9303 | 0.9002 | **0.9469** | 0.9033 | 0.9032 | 0.9339 | *0.9465* |
| | | MSE↓ | 0.0487 | 0.1693 | 0.4559 | *0.0144* | 0.0541 | 0.0542 | 0.0146 | **0.0113** |
| | | FID↓ | 323.4453 | 116.1988 | **91.3250** | 141.8722 | 147.9126 | 154.3030 | 133.0782 | *106.7131* |
| *Clinical* | 4 | PSNR↑ | 7.39 | 14.49 | 27.34 | 31.50 | 28.93 | 30.33 | *31.58* | **32.10** |
| | | SSIM↑ | 0.7589 | 0.8265 | 0.9349 | *0.9600* | 0.9252 | 0.9397 | 0.9541 | **0.9628** |
| | | MSE↓ | 0.1825 | 0.0573 | 0.0028 | *0.0004* | 0.0008 | 0.0006 | 0.0004 | **0.0003** |
| | | FID↓ | 327.5036 | 176.1845 | 132.2777 | *106.3749* | 163.8730 | 158.7533 | 107.9869 | **89.5350** |
| | 10 | PSNR↑ | 7.59 | 13.89 | 29.60 | *31.29* | 26.97 | 28.41 | 29.07 | **31.31** |
| | | SSIM↑ | 0.7588 | 0.8190 | 0.9362 | *0.9517* | 0.9064 | 0.9220 | 0.9278 | **0.9531** |
| | | MSE↓ | 0.1742 | 0.0842 | *0.0004* | 0.0004 | 0.0007 | 0.0006 | 0.0005 | **0.0004** |
| | | FID↓ | 309.4201 | 189.7446 | 136.3472 | *111.8106* | 164.3192 | 180.6273 | 126.5476 | **98.3469** |

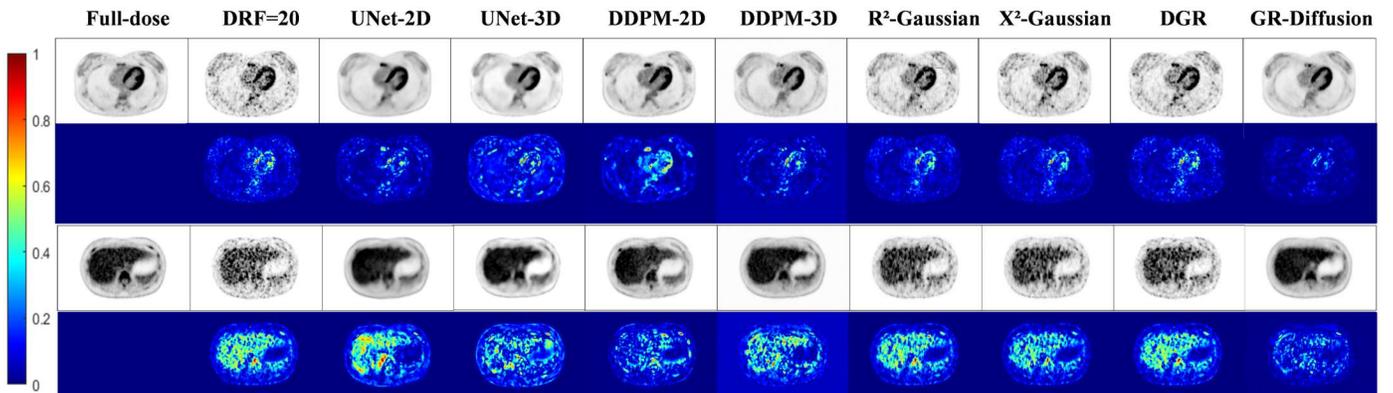

Fig. 6. Reconstruction results on the *UDPET* public dataset at a DRF value of 20. Two representative axial slices at different anatomical levels are shown. From left to right: Full-dose, low-dose, reconstruction by UNet-2D, UNet-3D, DDPM-2D, DDPM-3D, $R^2$-Gaussian, $X^2$-Gaussian, DGR, and GR-Diffusion. The corresponding error map with respect to the full-dose reference is presented below each reconstructed PET image.

*Comparison on Clinical Dataset:* To evaluate the reconstruction performance of the proposed method for clinical applications at DRF = 4 and DRF = 10 dose levels, we collected a clinical dataset and conducted a comparative assessment of multiple methods on it. In the lower part of Table I, we compare the deep learning-based methods including UNet-2D, UNet-3D, DDPM-2D and DDPM-3D and the Gaussian representation-based methods including $X^2$-Gaussian, $R^2$-Gaussian and DGR in terms of the image quality metrics PSNR, MSE and SSIM and the perceptual metric FID. Due to the more complex noise distributions and degradation models of Clinical low-dose data, overall metrics were inferior to those on the *UDPET* dataset. GR-Diffusion yielded the optimal performance across both dose levels. In terms of image quality, UNet-2D and UNet-3D achieved a PSNR of approximately 7 dB and an SSIM of around 0.75. This is because the UNet-based methods struggle to perform accurate reconstruction and tend to generate blurry, indistinct outputs when confronted with input contaminated by complex noise. In terms of perceptual, GR-Diffusion delivered leading performance, achieving FID values of 89.5350 and 98.3469. This demonstrates that the method of GR-Diffusion fusing accurate quantitative information with geometry-related information attains superior capability at the visual perceptual level.

Fig. 7 illustrates the performance of different methods across various ROIs in clinical reconstruction, where the results from top to bottom correspond to the reconstructed output of ROI 1, ROI 2 and ROI 3, respectively. ROI 1 corresponds to the reconstructed imaging of the cardiac region in patients. Severe noise in low-dose scans leads to a significant loss of cardiac morphological details. For UNet-2D and DDPM-2D, both of which perform reconstruction based on axial 2D imaging, the resultant outputs suffer from poor interslice consistency, making it difficult to form a continuous and complete 3D volume. In the case of $X^2$-Gaussian and $R^2$-Gaussian, the excessive degradation of low-dose images results in substantial residual noise in their reconstructed results, with the cardiac morphology being completely distorted. As for UNet-3D, DDPM-3D and DGR, although these methods avoid 3D consistency distortion and suppress a large amount of noise, they yield relatively low semi-quantitative values for the heart. This phenomenon may be attributed to the misclassification of critical quantitative information as noise during the reconstruction process. In contrast, GR-Diffusion preserves both 3D consistency and accurate quantitative information without any loss. ROI 2 corresponds to the liver region of the patient, and the reconstructions by GR-Diffusion are smooth and homogeneous with no excessive noise, achieving the best performance among all the methods. For ROI 3, a noisy spot with an abnormally high quantitative value is present in the low-dose images, which can severely interfere with the model judgment, causing the noise to be mistakenly identified as a lesion requiring reconstruction and even further amplified. Consequently, all other methods reconstruct a spurious lesion at this location. In contrast, GR-Diffusion is not perturbed by such noise and still performs accurate and faithful reconstruction.

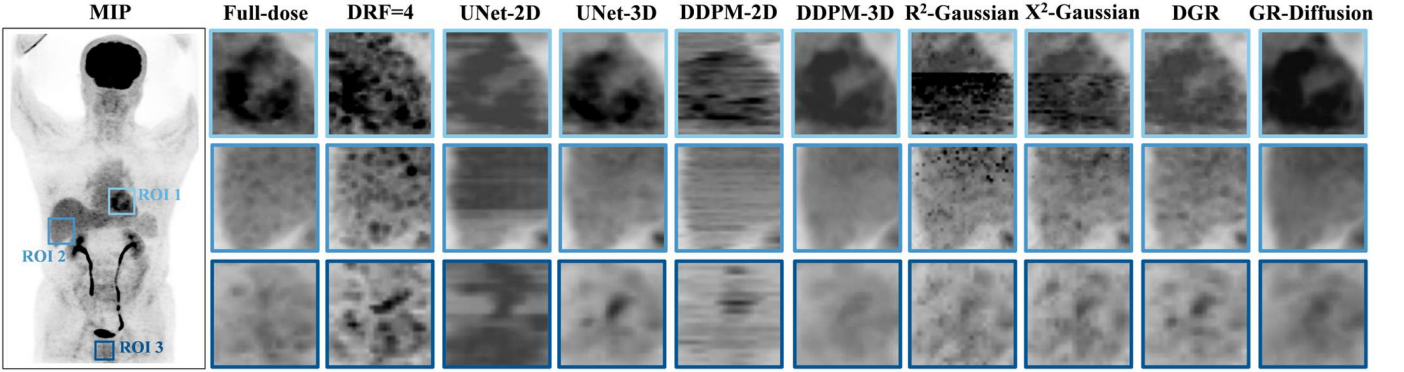

**Fig. 7.** Reconstruction results for ROIs on the *Clinical* dataset at DRF=4. From left to right: Full-dose, low-dose, UNet-2D, UNet-3D, DDPM-2D, DDPM-3D, $R^2$-Gaussian, $X^2$-Gaussian, DGR and GR-Diffusion. From top to bottom are the reconstructed results of ROI 1, ROI 2 and ROI 3, respectively.

### C. Ablation Study

To assess the contribution of individual components in our proposed method, we conduct ablation studies on the *UDPET* dataset, evaluating the impact of core components on reconstruction quality both quantitative and perceptual metrics.

***Effect of Gradient Guidance Terms:*** GR-Diffusion integrates geometric prior into the reverse diffusion process via coarse and fine-grained guidance terms. To systematically evaluate their contributions, we analyze four variants:

1) **Type I**: Excludes all guidance terms, utilizing only the low-dose PET image as condition.

2) **Type II**: Incorporates solely the coarse-grained guidance term: $-\omega_m \nabla_{x_t} \left\| G_{\sigma_m} * x_t - G_{\sigma_m} * x_{GR} \right\|_1$.

3) **Type III**: Employs only the fine-grained guidance term: $-\eta_n \nabla_{x_t} \left\| x_t - x_{GR} \right\|_1$.

4) **Type IV (Ours)**: Combines both coarse and fine-grained guidance terms.

Here, $x_t$ denotes the noisy PET image predicted during reverse process, $x_{GR}$ represents the Gaussian geometric prior, and all guidance terms are applied within the timestep interval $[0.4T, 0.6T]$. This strategy balances between computational resources and guidance strength, thus avoiding the introduction of ineffective guidance at early phase or the lead detail degradation at late phase of reverse process.

Fig. 8 illustrates the comparative performance of the four configuration types in terms of image quality and visual perception metrics. To comprehensively evaluate the 3D reconstruction quality and geometric consistency of the models, assessments were conducted across three anatomical planes: the coronal, sagittal and axial views. Overall, the sagittal tended to yield lower PSNR and SSIM values, along with higher MSE and FID scores, compared to the coronal and axial. This is attributed to the presence of more fine-grained details and subtle textures in the sagittal view. Then, for the sagittal, type IV achieved the highest PSNR of 27.46 dB and SSIM of 0.9562, as well as the lowest MSE of 0.2455 and FID of 55.1250, followed by type II and type III, with type I performing the worst. These results indicate that in scenarios rich in intricate details, GR-Diffusion which incorporates both coarse and fine-grained guidance, delivers the best performance. It balances details preservation without inducing morphological degradation from over-emphasis on local features, while also maintaining global quantitative fidelity without sacrificing fine local structures.

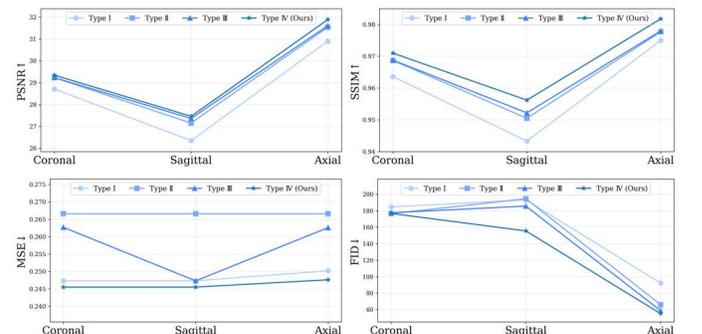

**Fig. 8.** Line charts of image-quality metrics (PSNR, SSIM and MSE) and perceptual metrics (FID) across three views (Coronal, Sagittal and Axial) in ablation study on guidance terms. The results demonstrate the superiority of the proposed methods.

Fig. 9 presents a comparative visual analysis of reconstruction results across the four gradient guidance configurations. All models successfully reconstruct full-dose PET images with anatomically correct morphology and well-defined edges. Specifically, type II, which employs only coarse-grained guidance, suffers from degradation in high-frequency struc-tures, with perceptible artifacts visible in the zoomed-in ROIs across the axial, sagittal, and coronal planes. In contrast, type III, relying solely on fine-grained guidance, leads to structural distortion due to over-emphasis on the prior, which is also observable in the enlarged ROIs of Fig. 9. The dual-guidance configuration (type IV) balances coarse and fine-grained gradient signals, thereby preventing both over-fitting to local details and over-regularization of global structures. This balanced interplay yields reconstructions that align most closely with the full-dose reference in terms of overall morphology as well as fine-structural fidelity.

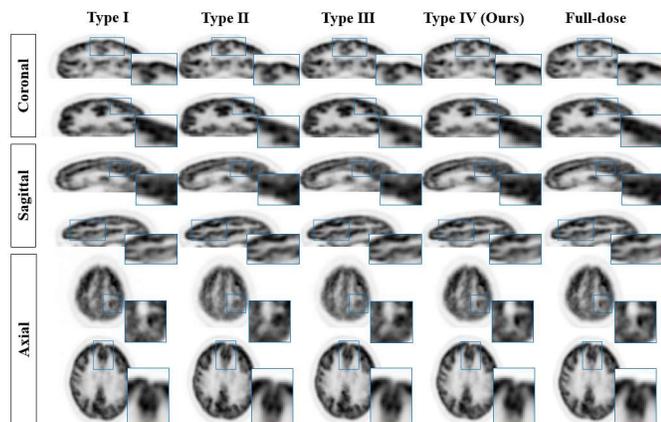

**Fig. 9.** Visual comparison of reconstruction results across three views (Coronal, Sagittal and Axial) in ablation study on guidance terms. Zoomed-in ROIs highlight regions with visible differences in image quality among the competing methods.

## IV. CONCLUSION

In this paper, we presented a GR-Diffusion framework that synergistically integrates 3D GR with a diffusion model for low-dose 3D whole-body PET reconstruction. GR-Diffusion leverages an explicit geometric prior derived from sinogram data via a discretized Gaussian encoding strategy to guide a generative diffusion process in a hierarchical manner. This approach effectively bridges the gap between physics-based reconstruction and data-driven generative refinement, enabling robust recovery of anatomical structures and metabolic details even under high noise and extreme low-dose conditions. Extensive experiments on the public *UDPET* and *Clinical* dataset demonstrated that GR-Diffusion consistently outperforms several state-of-the-art methods across multiple DRFs. Future work will focus on extending GR-Diffusion to dynamic and multi-tracer PET imaging scenarios and integrating additional physiological constraints for further improvement of clinical relevance and reliability.

## REFERENCES

[1] E. M. Rohren, T. G. Turkington, and R. E. Coleman, "Clinical applications of PET in oncology," *Radiology*, vol. 231, no. 2, pp. 305–332, 2004.

[2] M. Schwaiger, S. Ziegler, and S. G. Nekolla, "PET/CT: challenge for nuclear cardiology," *Journal of Nuclear Medicine*, vol. 46, no. 10, pp. 1664–1678, 2005.

[3] C. M. Clark, M. J. Pontecorvo, T. G. Beach, *et al.*, "Cerebral PET with florbetapir compared with neuropathology at autopsy for detection of neuritic amyloid-β plaques: a prospective cohort study," *Lancet Neurology*, vol. 11, no. 8, pp. 669–678, 2012.

[4] F. Hashimoto, Y. Onishi, K. Ote, *et al.*, "Deep learning-based PET image denoising and reconstruction: a review," Radiological physics and technology, vol. 17, no. 1, pp. 24-46, 2024.

[5] Y. Wang, G. Ma, L. An, *et al.*, "Semisupervised tripled dictionary learning for standard-dose PET image prediction using low-dose PET and multimodal MRI," *IEEE Transactions on Biomedical Engineering*, vol. 64, no. 3, pp. 569-579, 2016.

[6] L. Zhou, J. D. Schaefferkoetter, I. W. K. Tham, *et al.*, "Supervised learning with cyclegan for low-dose FDG PET image denoising," *Medical Image Analysis*, vol. 65, pp. 101770, 2020.

[7] B. Zhou, Y. J. Tsai, X. Chen, *et al.*, "MDPET: a unified motion correction and denoising adversarial network for low-dose gated PET," *IEEE Transactions on Medical Imaging*, vol. 40, no. 11, pp. 3154-3164, 2021.

[8] L. A. Feldkamp, "Practical cone beam algorithm," *J. Microsc.*, vol. 185, pp. 67-75, 1997.

[9] H. M. Hudson, R. S. Larkin, "Accelerated image reconstruction using ordered subsets of projection data," *IEEE Transactions on Medical Imaging*, vol. 13, no. 4, pp. 601-609, 1994.

[10] M. J. Ehrhardt, P. Markiewicz, M. Liljeroth, *et al.*, "PET reconstruction with an anatomical MRI prior using parallel level sets," *IEEE Transactions on Medical Imaging*, vol. 35, no. 9, pp. 2189-2199, 2016.

[11] B. Huang, X. Liu, L. Fang, *et al.*, "Diffusion transformer model with compact prior for low-dose PET reconstruction," *Physics in Medicine & Biology*, vol. 70, no. 4, pp. 045015, 2025.

[12] F. Hashimoto, Y. Onishi, K. Ote, *et al.*, "Deep learning-based PET image denoising and reconstruction: a review," Radiological physics and technology, vol. 17, no. 1, pp. 24-46, 2024.

[13] R. Guo, J. Wang, Y. Miao, *et al.*, "3D full-dose brain-PET volume recovery from low-dose data through deep learning: quantitative assessment and clinical evaluation," *European Radiology*, vol. 35, no. 3, pp. 1133-1145, 2025.

[14] Z. Peng, F. Zhang, J. Sun, *et al.*, "Preliminary deep learning-based low dose whole body PET denoising incorporating CT information," *2022 IEEE Nuclear Science Symposium and Medical Imaging Conference (NSS/MIC)*, pp. 1-2, 2022.

[15] M. Fu, M. Fang, B. Liao, *et al.*, "Low-count PET image reconstruction with generalized sparsity priors via unrolled deep networks," *IEEE Journal of Biomedical and Health Informatics*, 2025.

[16] O. Ronneberger, P. Fischer, T. Brox, "U-Net: Convolutional networks for biomedical image segmentation," *International Conference on Medical Image Computing and Computer-Assisted Intervention*, pp. 234-241, 2015.

[17] G. Chen, S. Liu, W. Ding, *et al.*, "A total-body ultralow-dose PET reconstruction method via image space shuffle U-Net and body sampling," *IEEE Transactions on Radiation and Plasma Medical Sciences*, vol. 8, no. 4, pp. 357-365, 2023.

[18] K. Kaviani, A. Sanaat, M. Mokri, *et al.*, "Image reconstruction using UNET-transformer network for fast and low-dose PET scans," *Computerized Medical Imaging and Graphics*, vol. 110, pp. 102315, 2023.

[19] I. J. Goodfellow, J. Pouget-Abadie, M. Mirza, *et al.*, "Generative adversarial nets," *Advances in Neural Information Processing Systems*, vol. 27, 2014.

[20] J. Cui, Y. Wang, L. Zhou, *et al.*, "3D point-based multi-modal context clusters GAN for low-dose PET image denoising," *IEEE Transactions on Circuits and Systems for Video Technology*, vol. 34, no. 10, pp. 9400-9413, 2024.

[21] J. Ho, A. Jain, P. Abbeel, "Denoising diffusion probabilistic models," *Advances in Neural Information Processing Systems*, vol. 33, pp. 6840-6851, 2020.

[22] Y. Song, J. Sohl-Dickstein, D. P. Kingma, *et al.*, "Score-based generative modeling through stochastic differential equations," *International Conference on Learning Representations*, 2021.

[23] J. Cui, X. Zeng, P. Zeng, *et al.*, "MGTP: Multi-granularity textual prompts for low-dose brain PET image denoising via adversarial diffusion model," *IEEE Journal of Biomedical and Health Informatics*, vol. 30, no. 1, pp. 448-458, 2026.

[24] K. Gong, F. Hashimoto, "PET image reconstruction using deep diffusion image prior," *IEEE Transactions on Medical Imaging*, vol. 38, no.7, pp. 1655-1665, 2018.

[25] B. D. O. Anderson, "Reverse-time diffusion equation models," *Stochastic Processes and their Applications*, vol. 12, no. 3, pp. 313-326, 1982.

[26] F. Djeumou, T. J. Lew, N. Ding, *et al.*, "One model to drift them all: Physics-informed conditional diffusion model for driving at the limits," *8th Annual Conference on Robot Learning*, 2024.

[27] B. Kerbl, G. Kopanas, T. Leimkühler, *et al*., "3D Gaussian splatting for real-time radiance field rendering," *ACM Trans*. Graph., vol. 42, no. 4, pp. 139:1-139:14, 2023.

[28] H. Nguyen, A. Le, B. R. Li, *et al*., "From coarse to fine: Learnable discrete wavelet transforms for efficient 3D Gaussian splatting," *Proceedings of the IEEE/CVF International Conference on Computer Vision*, 2025, pp. 3139-3148.

[29] C. Xu, Z. Jin, C. Shen, *et al*., "3D Gaussian adaptive reconstruction for Fourier light-field microscopy," *Advanced Imaging*, vol. 2, no. 5, pp. 055001, 2025.

[30] Y. Cai, H. Zhang, K. Zhang, *et al*., "Baking Gaussian splatting into diffusion denoiser for fast and scalable single-stage image-to-3d generation and reconstruction," *Proceedings of the IEEE/CVF International Conference on Computer Vision*, pp. 25062-25072, 2025.

[31] Y. Cheng, Y. Cai, Y. Zhang, "DenoiseGS: Gaussian reconstruction model for burst denoising," *arXiv preprint arXiv*:2511.22939, 2025.

[32] S. Jecklin, A. Massalimova, R. Zha, *et al*., "Intraoperative 3D reconstruction from sparse arbitrarily posed real X-rays," *Scientific Reports*, pp. 43973, 2025.

[33] F. Wang, J. Tao, J. Wu, *et al*., "X-Field: A physically informed representation for 3D X-ray reconstruction," *The Thirty-ninth Annual Conference on Neural Information Processing Systems*, 2025.

[34] Y. Cai, Y. Liang, J. Wang, *et al*., "Radiative Gaussian splatting for efficient X-ray novel view synthesis," *European Conference on Computer Vision, Cham: Springer Nature Switzerland*, 2024, pp. 283-299.

[35] Y. Lin, *et al*., "Learning 3D Gaussians for extremely sparse-view cone-beam CT reconstruction," *International Conference on Medical Image Computing and Computer-Assisted Intervention*, 2024.

[36] R. Zha, T. J. Lin, Y. Cai, *et al*., "R²-Gaussian: Rectifying radiative Gaussian splatting for tomographic reconstruction," *Advances in Neural Information Processing Systems*, vol. 37, pp. 44907-44934, 2024.

[37] Y. Yuluo, Y. Ma, K. Shen, *et al*., " Graph-based radiative Gaussian splatting for sparse-view CT reconstruction," *The Thirteenth International Conference on Learning Representations*, 2025.

[38] Y. Huang, I. Singh, T. Joyce, *et al*., "DIGS: Dynamic CBCT reconstruction using deformation-informed 4D Gaussian splatting and a low-rank free-form deformation model," *International Conference on Medical Image Computing and Computer-Assisted Intervention*, pp. 131-141, 2025.1

[39] W. Yu, Y. Cai, R. Zha, *et al*., "X²-Gaussian: 4D radiative Gaussian splatting for continuous-time tomographic reconstruction," *Proceedings of the IEEE/CVF International Conference on Computer Vision (ICCV)*, 2025.

[40] T. Peng, R. Zha, Z. Li, *et al*., "Three-dimensional MRI reconstruction with Gaussian representations: Tackling the undersampling problem," *IEEE Transactions on Medical Imaging*, 2025.

[41] S. Wu, Y. Lu, Y. Guo, *et al*., "Discretized Gaussian representation for tomographic reconstruction," *Proceedings of the IEEE/CVF International Conference on Computer Vision*, pp. 25073-25082, 2025.

[42] H. Qu, X. Wang, G. Zhang, *et al*., "GEM: 3D Gaussian splatting for efficient and accurate Cryo-EM reconstruction," *arXiv preprint arXiv*:2509.25075, 2025.

[43] F. Duelmer, J. Klaushofer, M. Wysocki, *et al*., "UltraG-Ray: Physics-based Gaussian ray casting for novel ultrasound view synthesis," *Medical Imaging with Deep Learning*, 2026.

[44] Y. Yang, W. Cai, D. Yang, *et al*., "UltraGS: Gaussian splatting for Ultrasound novel view synthesis," *arXiv preprint arXiv*:2511.07743, 2025.

[45] J. Xie, H. C. Shao, Y. Zhang, "Time-resolved dynamic CBCT reconstruction using prior-model-free spatiotemporal Gaussian representation (PMF-STGR)," *Physics in Medicine & Biology*, vol. 70, no. 16, pp. 165011, 2025.

[46] W. Zhang, H. Zhu, D. Wu, *et al*., "WIPES: Wavelet-based visual primitives," *Proceedings of the IEEE/CVF International Conference on Computer Vision*, 2025, pp. 27338-27347.

[47] B. Yu, S. Ozdemir, Y. Dong, *et al.*, "Robust whole-body PET image denoising using 3D diffusion models: evaluation across various scanners, tracers, and dose levels," *European Journal of Nuclear Medicine and Molecular Imaging*, vol. 52, no. 7, pp. 2549-2562, 2025.